\documentclass{article}
\usepackage{iclr2025_conference,times}
\iclrfinalcopy
\usepackage{amsmath}
\usepackage{amssymb}
\usepackage{graphicx}
\usepackage{booktabs}
\usepackage{algorithm}
\usepackage{algpseudocode}
\usepackage{hyperref}
\usepackage{url}

\graphicspath{{figures/}}

\newcommand{\grow}{\textsc{Grow}}
\newcommand{\prune}{\textsc{Prune}}
\newcommand{\branch}{\textsc{Branch}}
\newcommand{\base}{\textsc{CoT}}

\title{Recursive Agentic Reasoning}

\author{
Shengxin Zhang \\
Google \\
\texttt{lezhend@gmail.com}
\And
Xiaomin Wu \\
University of Maryland \\
\texttt{fred.xiaomin.wu@gmail.com}
\And
Xiyang Wu \\
University of Maryland \\
\texttt{wuxiyang@umd.edu}
\And
Jing Xie \\
University of Maryland \\
\texttt{jingxie@terpmail.umd.edu}
}

\begin{document}

\maketitle

\begin{abstract}
Test-time reasoning methods are usually studied in isolation: iterative refinement, decomposition, and repeated sampling each arrive with their own benchmarks, models, and graders, so their gains are hard to compare. We recast them as \emph{recursion operators} over an agent's reasoning trace --- \grow{} (deepen one path), \prune{} (decompose and recompose), and \branch{} (sample alternatives and select) --- and evaluate them against a single-pass chain-of-thought baseline in one shared harness with identical prompts, token budgets, and grading code. The study spans five benchmarks and three frontier models, yielding 14 model $\times$ benchmark cells, 49{,}327 graded items, and 151{,}876 model calls. Under a paired protocol that scores every operator on exactly the items resolved by all operators, \branch{} improves accuracy in all 14 cells (mean $+5.98$ points) and is the strict best operator in 12; \grow{} averages $+2.18$ and is negative in two cells, while \prune{} averages $+0.94$. The central analytical result is that \branch{}'s advantage is driven not only by marginalization over reasoning paths, but also by truncation recovery. Its gain strongly tracks the baseline's rate of empty, budget-exhausted outputs ($r = 0.72$), indicating that repeated sampling often recovers answers that a single pass never emits. This finding weakens the routing hypothesis that motivated the study: at this operator granularity, one method dominates nearly everywhere. We also show that unpaired evaluation and scoring infrastructure failures as model errors are large enough to invert conclusions, and we recommend paired scoring as standard practice for comparative test-time-compute studies.
\end{abstract}

\section{Introduction}

A language model given one attempt at a hard problem produces one trace and one answer. Almost every recent advance in test-time reasoning can be read as a way of spending additional inference compute to avoid being stuck with that single trace. The model can extend the trace and revise its own answer \citep{madaan2023selfrefine,shinn2023reflexion,shi2025ssr}; it can cut the problem into sub-problems and solve them in sequence \citep{zhou2023leasttomost,khot2023decomposed}; or it can sample many independent traces and aggregate them \citep{wang2023selfconsistency,brown2024monkeys,chang2025step}. Recent work has pushed this space further with learned search controllers \citep{li2025pgts}, adaptive compute allocation \citep{wang2025every,bilal2026adaptive}, and explicitly recursive agents that delegate to copies of themselves \citep{yang2026recursive,gandhi2026rao}. These are different shapes of computation, and the empirical literature still largely treats them as different research programs.

That separation has a practical cost. Each method is typically introduced on the benchmarks and base models that suit it, with its own answer-extraction and grading code. A practitioner deciding how to spend a fixed inference budget therefore cannot answer a basic question: at equal compute, which of these should I use, and does the answer depend on my model or my task? Reported gains are not comparable, because nothing except the method is held fixed. This concern has only become sharper as recent framework papers argue that test-time scaling should be evaluated as a full inference system, with protocol-matched compute accounting and reproducibility requirements rather than a single scalar budget \citep{hariri2026tts}.

\begin{figure}[t]
\centering
\includegraphics[width=\textwidth]{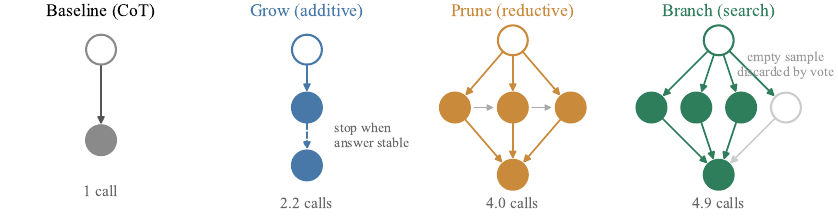}
\caption{The four inference-time computation graphs compared in this paper. Hollow nodes are the problem, filled nodes are model calls, and the bottom node is the returned answer. \base{} issues one call. \grow{} extends a single path and halts when the extracted answer stops changing between rounds. \prune{} decomposes into ordered sub-questions, solves each with the previous answers in context, and recomposes. \branch{} samples $N = 5$ independent solutions and selects by majority vote over normalized answers, which also discards samples that returned nothing. Call counts are means measured over all runs.}
\label{fig:operators}
\end{figure}

This paper takes the position that the three families are better understood as three \emph{operators} in a single space --- recursion over an agent's reasoning trace --- and that the interesting question is empirical and comparative. We define \grow{} (additive recursion along one path), \prune{} (reductive recursion by decomposition), and \branch{} (recursion by search over sampled alternatives), implement them over a shared primitive, and run all three against a single-pass chain-of-thought baseline \citep{wei2022chain} on five benchmarks and three frontier models. Unlike recent work that trains recursive or policy-guided inference systems directly \citep{li2025pgts,yang2026recursive,gandhi2026rao}, we ask a narrower but more immediately deployable question: if the base model is held fixed, which operator buys the most accuracy and why? Figure~\ref{fig:operators} shows the four computation graphs.

Holding everything but the operator fixed turns out to matter in ways we did not anticipate. Two methodological hazards, both invisible in a conventional results table, were individually large enough to reverse conclusions in our study. The first is unpaired evaluation: when long-running jobs against a shared inference endpoint fail on different items for different operators, comparing per-operator accuracy over whatever each run happened to complete silently compares methods on different problem sets. In our data this manufactured a spurious 3-point regression. The second is treating infrastructure failures as model errors, which penalizes exactly the operators that issue the most API calls --- that is, the most expensive and often the best ones. We therefore adopt a paired protocol throughout: within each model $\times$ benchmark cell, every operator is scored on precisely the set of items that all operators resolved, and unrecoverable transport failures are excluded rather than marked wrong.

Our contributions are:

\begin{itemize}
\item \textbf{A unified operator formulation} of additive, reductive, and search-based test-time recursion over a reasoning trace, built on a shared solving primitive that includes an explicit finalization step for budget-exhausted reasoning models (Section~\ref{sec:method}).

\item \textbf{A controlled 14-cell comparison} --- 5 benchmarks $\times$ 3 models $\times$ 4 methods, 49{,}327 graded items, 151{,}876 model calls --- with identical prompts, budgets, and graders, using a paired scoring protocol (Sections~\ref{sec:setup}--\ref{sec:results}). \branch{} improves accuracy in all 14 cells and wins 12; \grow{} and \prune{} are inconsistent.

\item \textbf{A mechanistic account of why \branch{} wins} that differs from the standard explanation. Its gain tracks the baseline's rate of empty, budget-exhausted outputs at $r = 0.72$, and it roughly halves that rate; the benefit is substantially recovery of a degenerate failure mode rather than marginalization over reasoning paths (Section~\ref{sec:mechanism}).

\item \textbf{Two negative results.} Adaptive operator routing is not supported by the data, because a single operator dominates; and we document how unpaired scoring inverted one of our own findings (Sections~\ref{sec:results} and \ref{sec:discussion}).
\end{itemize}

The rest of the paper is organized as follows. Section~\ref{sec:method} formalizes the three operators and the shared solving primitive. Section~\ref{sec:setup} describes the benchmarks, models, and paired evaluation protocol. Section~\ref{sec:results} gives the main comparative result, explains the mechanism behind \branch{}'s gains, and analyzes cost and cross-model effects. Section~\ref{sec:discussion} situates the findings relative to the prior literature and makes the paper's negative results explicit.

\section{Related Work}

\paragraph{Prompting and decomposition.} Chain-of-thought prompting established that eliciting intermediate steps improves reasoning \citep{wei2022chain}. Least-to-most prompting \citep{zhou2023leasttomost} and Decomposed Prompting \citep{khot2023decomposed} break a problem into sub-problems solved in sequence, each conditioned on earlier answers. Our \prune{} operator is a direct descendant of this line, with the decomposition produced by the same model that solves the sub-questions and an adaptive gate that falls back to a single direct solve when the model proposes only one sub-question.

\paragraph{Iterative refinement.} Self-Refine \citep{madaan2023selfrefine} and Reflexion \citep{shinn2023reflexion} improve an output by feeding the model its own critique. More recent work pushes refinement to finer granularity: Socratic Self-Refine decomposes a reasoning trace into verifiable sub-question/sub-answer pairs, estimates step-level confidence, and selectively revises the weakest step \citep{shi2025ssr}. Our \grow{} operator is a deliberately minimal member of this family: it re-solves with the previous attempt in context and uses \emph{answer stability} --- two consecutive rounds extracting the same normalized answer --- as its halting signal, which requires no separate critic and no verbal feedback channel.

\paragraph{Sampling and search.} Self-consistency samples multiple chains and takes a majority vote \citep{wang2023selfconsistency}. Tree of Thoughts \citep{yao2023tot} and Graph of Thoughts \citep{besta2024got} expand and prune partial reasoning states with lookahead and backtracking. Recent work spends branching compute more deliberately: Policy-Guided Tree Search learns when to expand, branch, backtrack, or terminate \citep{li2025pgts}, while step-level hybrid test-time scaling interleaves verifier-guided self-refinement with Best-of-$N$ and MCTS inside a single inference routine \citep{chang2025step}. Our \branch{} operator sits at the simple end of this spectrum: it is flat parallel sampling with an unweighted vote, and it is important to state plainly that as an algorithm it is self-consistency. We name it \branch{} for symmetry within the operator space, not to claim novelty; Section~\ref{sec:discussion} discusses what does and does not distinguish our findings from that prior work.

\paragraph{Recursive agentic systems.} Recent 2026 work turns recursion itself into the object of study. Recursive Models for Long-Horizon Reasoning formalize self-recursive agents as a route around bounded context and analyze their computational power \citep{yang2026recursive}, while Recursive Agent Optimization trains agents end-to-end to decide when and how to delegate to recursive copies of themselves \citep{gandhi2026rao}. These works are complementary to ours: they study learned recursive policies directly, whereas we compare lightweight black-box recursion operators that can be layered onto existing frontier models without additional training.

\paragraph{Test-time compute scaling.} \citet{brown2024monkeys} show that coverage --- the fraction of problems solved by at least one sample --- scales log-linearly with sample count over four orders of magnitude, while noting that selection methods such as majority voting plateau beyond several hundred samples. \citet{snell2024scaling} show that the best way to spend test-time compute depends on problem difficulty, and that adaptively allocating it beats a fixed best-of-$N$ policy. Recent work sharpens both allocation and evaluation: Every Rollout Counts studies optimal rollout allocation during search and argues for allocating budget at the reasoning-direction level rather than the solution level \citep{wang2025every}; \citet{bilal2026adaptive} propose verifier-guided adaptive allocation across tools, exploration parameters, and iteration budgets; and \citet{hariri2026tts} argue that test-time scaling should be reported as whole inference systems with protocol-matched compute accounting and reproducibility artifacts. This line of work motivates the routing hypothesis we set out to test; our data does not support it at the operator granularity, which we take to be an informative negative result rather than a contradiction, since we vary the operator rather than the sample budget. Verifier-guided selection \citep{cobbe2021verifiers,lightman2024verify} is also where much of the recent progress sits, and it is the main axis we do \emph{not} explore: all of our selection is unweighted voting, which we return to as the clearest limitation of this study.

\section{Method}
\label{sec:method}

\subsection{Agentic recursion over a reasoning trace}

Let $x$ be a problem and let a model call $M(p)$ map a prompt to a completion. A single-pass baseline computes $y = \mathrm{extract}(M(x))$. We define a recursion operator as a policy that composes several such calls into one answer, and we characterize each operator by the computation graph it induces (Figure~\ref{fig:operators}) and by its halting rule. We call the overall procedure \emph{agentic} because the model is allowed to issue multiple calls conditioned on its own intermediate products rather than committing to a single trace.

\subsection{Shared solving primitive}

All operators are built on one function, $\mathrm{solve}(x, c)$, which issues a single call with optional prior context $c$ and returns an extracted answer. It contains one piece of machinery that proved essential and that we consider a contribution in its own right. Contemporary reasoning models emit a hidden deliberation stream before their user-visible content; if the token budget is exhausted mid-deliberation, the call returns successfully with \emph{empty} content and a length stop reason. Treating this as a wrong answer --- the default in most harnesses --- discards an item on which the model may have reasoned correctly for thousands of tokens. When $\mathrm{solve}$ detects this condition it re-prompts the model with its own truncated reasoning and asks it to commit to a final answer. Section~\ref{sec:mechanism} shows that this failure mode is not a corner case: it accounts for a majority of baseline items on our hardest benchmark.

\subsection{Three recursion operators}

\paragraph{\grow{}: additive recursion.} \grow{} deepens a single path. Round $t$ calls $\mathrm{solve}(x, y_{t-1})$, placing the previous attempt in context. It halts when $y_t = y_{t-1}$ under answer normalization, or at a cap of three rounds. The premise is that when a first attempt is under-determined --- too little information, too short a chain --- another pass with the attempt visible will improve it, and that agreement between consecutive rounds is a usable proxy for convergence.

\paragraph{\prune{}: reductive recursion.} \prune{} reduces an over-complex problem. One call decomposes $x$ into an ordered list of sub-questions $q_1, \dots, q_k$; each $q_i$ is solved with the answers to $q_{<i}$ in context, which matters for multi-hop problems where later hops depend on earlier ones; a final call composes the sub-answers into an answer to $x$. If the model proposes $k = 1$, the operator degrades to a single direct solve rather than paying the decomposition overhead.

\paragraph{\branch{}: recursion by search.} \branch{} explores alternatives. It draws $N = 5$ independent solutions at temperature $0.7$, normalizes each extracted answer to a stable key, and returns the plurality key, recording the agreement ratio. Because empty completions normalize to a distinguished null key that never wins a vote unless every sample is empty, the vote structurally discards budget-exhausted samples.

\subsection{Shared control signals}

The three operators differ in graph shape but share the same style of control signal: \grow{} halts on answer stability across rounds, \prune{} adapts its width to the model's own proposed decomposition, and \branch{} records agreement across samples. All three signals are computed from extracted answers using the same normalizer, and none requires a trained verifier, a reward model, or access to ground truth. This is what makes the comparison in Section~\ref{sec:results} apples-to-apples: any accuracy difference between operators is attributable to the shape of the recursion, not to a difference in supervision.

\section{Experimental Setup}
\label{sec:setup}

\subsection{Models and benchmarks}

We evaluate three frontier models --- DeepSeek-V4-Pro, MiniMax-M3, and Qwen3.6-plus --- accessed through a single internal OpenAI-compatible evaluation proxy, so that routing, retry policy, and token accounting are identical across models.\footnote{Model identifiers as sent to the proxy: \texttt{deepseek/deepseek-v4-pro}, \texttt{api\_minimax/MiniMax-M3}, \texttt{api\_ali/qwen3.6-plus}. All three are used as text-only endpoints.} All three are reasoning models that emit a hidden deliberation stream, which is what makes the truncation behavior of Section~\ref{sec:mechanism} relevant.

Benchmarks were selected for \emph{headroom}: we required that a strong model score well below ceiling, since a saturated benchmark cannot show a difference between test-time methods. An initial round on saturated mathematics sets was discarded for exactly this reason. Table~\ref{tab:benchmarks} lists the final selection, which spans grounded multi-hop retrieval, expert-level closed-ended questions, general reasoning, graduate domain knowledge, and olympiad mathematics.

\begin{table}[t]
\centering
\caption{Benchmarks. $n$ is the number of items evaluated per operator per model. MuSiQue supplies all supporting and distractor paragraphs in context, so it measures multi-hop \emph{reasoning} rather than retrieval. HLE is restricted to its text-only subset. $^{\dagger}$BBEH replaces each task of BIG-Bench Hard \citep{suzgun2023bbh} with a harder counterpart probing the same skill.}
\label{tab:benchmarks}
\small
\setlength{\tabcolsep}{5pt}
\begin{tabular}{llrl}
\toprule
Benchmark & Capability probed & $n$ & Metric \\
\midrule
MuSiQue \citep{trivedi2022musique} & Multi-hop composition   & 2{,}417 & EM / F1 \\
HLE \citep{phan2026hle}            & Expert-level academic   & 2{,}158 & Normalized EM \\
BBEH \citep{kazemi2025bbeh}        & General many-hop$^{\dagger}$ & 200 & Normalized EM \\
SuperGPQA \citep{du2025supergpqa}  & Graduate knowledge      & 300     & Multiple choice \\
Omni-MATH \citep{gao2025omnimath}  & Olympiad mathematics    & 300     & Normalized EM \\
\bottomrule
\end{tabular}
\end{table}

Grading is deliberately conservative: multiple-choice items are graded by letter extraction, and free-form items by symbolic and string normalization. For HLE this is a strict lower bound relative to the official LLM judge, so our absolute HLE numbers understate true accuracy. Because the same grader is applied to every operator within a cell, it does not bias the comparisons that this paper is about.

\subsection{Paired evaluation protocol}
\label{sec:protocol}

Runs of this length against a shared endpoint fail intermittently. Across the full study, 709 items terminated in unrecoverable transport failures --- read timeouts, connection resets, and provider rate limiting --- after seven retries with exponential backoff. These are not model errors, and how they are handled changes conclusions. We adopt three rules.

\textbf{Deduplicate by item.} The runner is resumable and appends retried items, so a raw result file can contain several records for the same item. Naive line-level accounting is badly wrong: one of our files reads as 36.3\% accurate by line count and 72.5\% after deduplication. We keep one record per item, preferring a successful record over a failed one.

\textbf{Exclude, do not penalize.} An item that never produced a completion is dropped rather than scored as incorrect. Scoring it wrong systematically penalizes operators that issue more calls per item, since their probability of hitting at least one failure is higher --- which is to say it penalizes \branch{} and \prune{} for being expensive.

\textbf{Score on the paired set.} Within each model $\times$ benchmark cell we intersect the resolved item sets of all four methods and report accuracy on that intersection. Every comparison in this paper is therefore between methods answering the identical questions. Table~\ref{tab:main} reports the resulting $n$ per cell; attrition is zero in six of the fourteen cells and below 1\% in four more, and Appendix~\ref{app:failures} gives the per-cell breakdown.

Section~\ref{sec:discussion} describes the finding that this protocol overturned.

\section{Results}
\label{sec:results}

\subsection{Overall comparison}

\begin{figure}[t]
\centering
\includegraphics[width=0.95\textwidth]{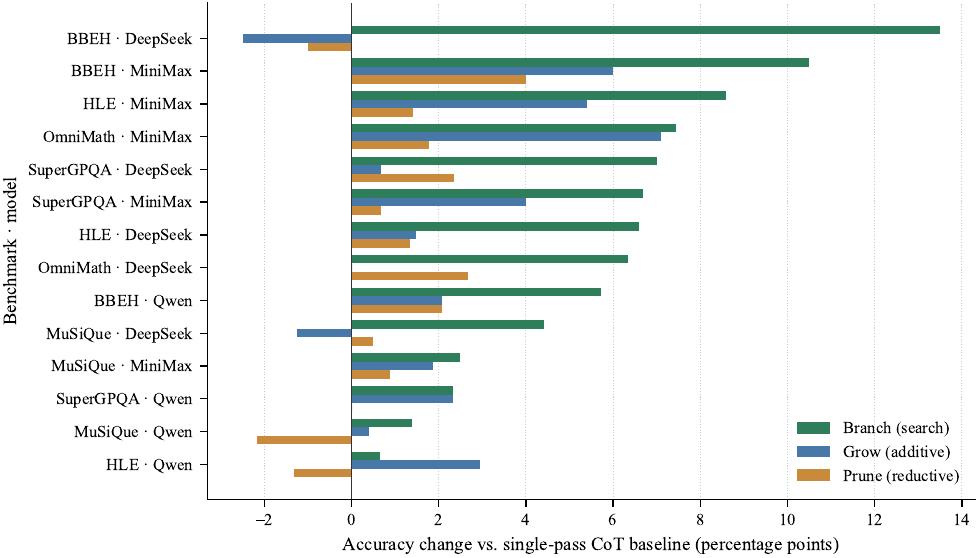}
\caption{Accuracy change relative to the single-pass \base{} baseline in each of the 14 model $\times$ benchmark cells, sorted by \branch{} gain. Bars right of zero are improvements. \branch{} is positive in every cell; \grow{} and \prune{} each go negative in two. All three operators are scored on the same paired item set as the baseline within a cell.}
\label{fig:gains}
\end{figure}

\begin{table}[t]
\centering
\caption{Accuracy (\%) by benchmark, model, and operator, with the change from the \base{} baseline in parentheses. $n$ is the paired item count (Section~\ref{sec:protocol}). Bold marks the best operator in each cell. Omni-MATH was not run on Qwen3.6-plus, leaving 14 of 15 possible cells.}
\label{tab:main}
\small
\setlength{\tabcolsep}{4pt}
\begin{tabular}{llrrrrr}
\toprule
Benchmark & Model & $n$ & \base{} & \grow{} & \prune{} & \branch{} \\
\midrule
MuSiQue & DeepSeek-V4-Pro & 2{,}417 & 67.94 & 66.69 \small{($-$1.25)} & 68.43 \small{($+$0.49)} & \textbf{72.36} \small{($+$4.42)} \\
MuSiQue & MiniMax-M3      & 2{,}398 & 73.35 & 75.23 \small{($+$1.88)} & 74.23 \small{($+$0.88)} & \textbf{75.85} \small{($+$2.50)} \\
MuSiQue & Qwen3.6-plus    & 2{,}395 & 77.08 & 77.49 \small{($+$0.41)} & 74.91 \small{($-$2.17)} & \textbf{78.46} \small{($+$1.38)} \\
\midrule
HLE & DeepSeek-V4-Pro & 2{,}152 & 13.34 & 14.82 \small{($+$1.48)} & 14.68 \small{($+$1.34)} & \textbf{19.93} \small{($+$6.59)} \\
HLE & MiniMax-M3      & 500     & 22.40 & 27.80 \small{($+$5.40)} & 23.80 \small{($+$1.40)} & \textbf{31.00} \small{($+$8.60)} \\
HLE & Qwen3.6-plus    & 305     & 16.39 & \textbf{19.34} \small{($+$2.95)} & 15.08 \small{($-$1.31)} & 17.05 \small{($+$0.66)} \\
\midrule
BBEH & DeepSeek-V4-Pro & 200 & 38.50 & 36.00 \small{($-$2.50)} & 37.50 \small{($-$1.00)} & \textbf{52.00} \small{($+$13.50)} \\
BBEH & MiniMax-M3      & 200 & 26.50 & 32.50 \small{($+$6.00)} & 30.50 \small{($+$4.00)} & \textbf{37.00} \small{($+$10.50)} \\
BBEH & Qwen3.6-plus    & 192 & 66.67 & 68.75 \small{($+$2.08)} & 68.75 \small{($+$2.08)} & \textbf{72.40} \small{($+$5.73)} \\
\midrule
SuperGPQA & DeepSeek-V4-Pro & 300 & 58.33 & 59.00 \small{($+$0.67)} & 60.67 \small{($+$2.34)} & \textbf{65.33} \small{($+$7.00)} \\
SuperGPQA & MiniMax-M3      & 299 & 58.53 & 62.54 \small{($+$4.01)} & 59.20 \small{($+$0.67)} & \textbf{65.22} \small{($+$6.69)} \\
SuperGPQA & Qwen3.6-plus    & 300 & 69.67 & \textbf{72.00} \small{($+$2.33)} & 69.67 \small{($+$0.00)} & \textbf{72.00} \small{($+$2.33)} \\
\midrule
Omni-MATH & DeepSeek-V4-Pro & 299 & 33.11 & 33.11 \small{($+$0.00)} & 35.79 \small{($+$2.68)} & \textbf{39.46} \small{($+$6.35)} \\
Omni-MATH & MiniMax-M3      & 282 & 30.14 & 37.23 \small{($+$7.09)} & 31.91 \small{($+$1.77)} & \textbf{37.59} \small{($+$7.45)} \\
\bottomrule
\end{tabular}
\end{table}

Figure~\ref{fig:gains} and Table~\ref{tab:main} give the main comparative result. \branch{} improves on the baseline in all 14 cells, with a mean gain of $+5.98$ points, a median of $+6.47$, and a range from $+0.66$ to $+13.50$. It is the strict best of the three operators in 12 cells, ties for best in one, and is beaten in one. \grow{} averages $+2.18$ and is positive in 11 of 14, but it is negative on DeepSeek-V4-Pro for both MuSiQue ($-1.25$) and BBEH ($-2.50$), so it cannot be recommended unconditionally. \prune{} averages $+0.94$ with a range of $-2.17$ to $+4.00$ --- an effect small enough that, at the sample sizes of our smaller cells, much of it is difficult to distinguish from noise.

The two cells \branch{} does not win are both Qwen3.6-plus, and they are instructive rather than contradictory. On SuperGPQA, \grow{} and \branch{} tie exactly at $+2.33$. On HLE, \grow{} leads with $+2.95$ against \branch{}'s $+0.66$, in the one cell with the heaviest attrition ($n = 305$ of 500 attempted). Qwen3.6-plus is also the only model in our study that never produced an empty completion, which Section~\ref{sec:mechanism} identifies as the precise condition under which \branch{}'s advantage should be expected to shrink.

\paragraph{The routing hypothesis is not supported.} This study was designed to test whether the best operator depends on the problem, which would motivate an adaptive router that selects an operator per item. The data argues against that hypothesis at this granularity. \branch{} is best in all five DeepSeek-V4-Pro cells and all five MiniMax-M3 cells; only Qwen3.6-plus splits, two to two, and one of those two is an exact tie. A router trained on these outcomes would learn a nearly constant policy. We regard that negative result as more informative than a router built anyway on the two cells that would support it.

\subsection{Why \branch{} wins: truncation recovery}
\label{sec:mechanism}

\begin{figure}[t]
\centering
\includegraphics[width=\textwidth]{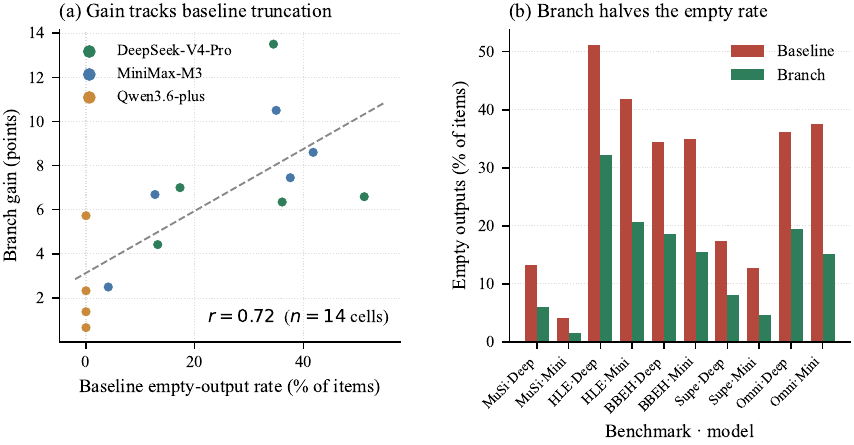}
\caption{(a) Per-cell \branch{} gain against the baseline's empty-output rate, with a least-squares fit; the association is strong ($r = 0.72$, $n = 14$). The four cells at zero on the horizontal axis are the Qwen3.6-plus cells, which never truncated, and they show the smallest gains. (b) Empty-output rate under the baseline and under \branch{} for the ten cells where truncation occurs at all. \branch{} roughly halves it everywhere.}
\label{fig:mechanism}
\end{figure}

Self-consistency is usually explained as marginalizing over reasoning paths: a hard problem admits many routes to one correct answer, so agreement across sampled routes is evidence for that answer \citep{wang2023selfconsistency}. Our data indicates that in this setting a different mechanism carries much of the gain.

The dominant baseline failure is not an incorrect answer but an absent one. Under a single pass, DeepSeek-V4-Pro returns empty content on 51.2\% of HLE items: the model exhausts a 16{,}000-token budget inside its hidden deliberation stream and never reaches user-visible content. The rate is 34.5\% on BBEH and 36.1\% on Omni-MATH. Sampling five times converts this from a near-coin-flip into a minority event, and because empty answers cannot win a plurality vote, they are discarded automatically.

Figure~\ref{fig:mechanism} makes the case quantitatively. Panel (a) shows that a cell's \branch{} gain is strongly associated with how often the baseline truncated ($r = 0.72$ over 14 cells). Panel (b) shows the reduction directly: on HLE with DeepSeek-V4-Pro the empty rate falls from 51.2\% to 32.2\%, on BBEH with MiniMax-M3 from 35.0\% to 15.5\%, and the rate drops in all ten cells where truncation occurs. The cleanest evidence is the negative case. Qwen3.6-plus never truncated on any benchmark, and it is exactly where \branch{}'s gains are smallest ($+0.66$, $+1.38$, $+2.33$, $+5.73$) and where the one non-tied loss to \grow{} occurs.

We are not claiming that marginalization contributes nothing --- Qwen3.6-plus still gains from \branch{} on three of four benchmarks with no truncation to recover. The claim is narrower: the two mechanisms are separable, they are conflated in aggregate accuracy numbers, and in a regime of long-reasoning models under finite budgets the recovery term is large. This has a practical consequence: a substantial fraction of the reported benefit of repeated sampling on such models may be obtainable far more cheaply by a targeted finalization pass on truncated generations, of the kind our $\mathrm{solve}$ primitive performs.

\subsection{Compute, efficiency, and cross-model comparison}

\begin{figure}[t]
\centering
\begin{minipage}[t]{0.47\textwidth}
\centering
\includegraphics[width=\textwidth]{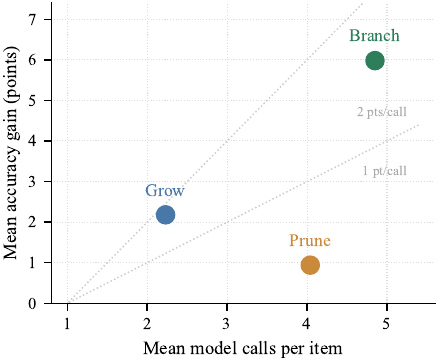}
\caption{Mean accuracy gain against mean model calls per item, averaged over all 14 cells. Dotted guides mark constant returns per call. \branch{} buys the most accuracy; \grow{} is marginally more efficient per call but has a far lower ceiling; \prune{} is dominated on both axes.}
\label{fig:cost}
\end{minipage}\hfill
\begin{minipage}[t]{0.47\textwidth}
\centering
\includegraphics[width=\textwidth]{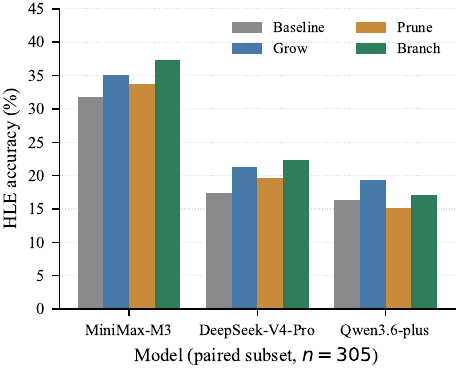}
\caption{HLE accuracy on the 305 items resolved by every model $\times$ operator run. Restricting to a shared item set changes the model ranking relative to the unpaired numbers in Table~\ref{tab:main}, because the 500-item subset attempted by two of the models is easier than the full split.}
\label{fig:hlefair}
\end{minipage}
\end{figure}

\paragraph{Cost.} The operators are not compute-comparable, and Figure~\ref{fig:cost} places them on the accuracy-versus-calls plane. \grow{} averages 2.23 calls per item, \prune{} 4.04, and \branch{} 4.85, against one for the baseline. Dividing mean gain by extra calls, \grow{} returns $+1.77$ points per additional call, \branch{} $+1.55$, and \prune{} $+0.31$. So \branch{}'s dominance is a dominance in \emph{achievable} accuracy rather than in efficiency: it wins because it can spend more usefully, not because each call is better spent. \prune{} is dominated on both axes and we see no budget at which we would recommend it.

The obvious inefficiency is that \branch{} uses a fixed $N = 5$ with no early stopping. Because the agreement ratio is already computed per item, halting once a majority is mathematically decided would cut a large share of the cost with no change in the returned answer. We did not implement this, and it is the first thing we would add.

\paragraph{Cross-model comparison requires the same items.} The full HLE split was run only on DeepSeek-V4-Pro; the other two models were run on a 500-item subset. The per-model HLE numbers in Table~\ref{tab:main} are therefore \emph{not} comparable across models, and the direction of the error is not obvious in advance. Figure~\ref{fig:hlefair} restricts all three models and all four operators to the 305 items every run resolved. On that shared set, DeepSeek-V4-Pro's baseline rises from 13.34\% to 17.38\%, because the 500-item subset is easier than the full split, and MiniMax-M3 leads at 31.80\%. Any cross-model claim from the unpaired table would have been wrong about both the gap and the ordering.

\section{Discussion}
\label{sec:discussion}

\subsection{What is and is not new}

\branch{} is self-consistency \citep{wang2023selfconsistency}. \grow{} is a stripped-down member of the Self-Refine family \citep{madaan2023selfrefine}, and \prune{} is least-to-most decomposition \citep{zhou2023leasttomost}. We claim novelty for none of the three as algorithms. The paper's contribution is instead the controlled comparison that the field's method-at-a-time convention makes unavailable, the finding that a large part of repeated sampling's benefit on long-reasoning models is truncation recovery rather than path marginalization, and a negative result on operator routing. Relative to \citet{brown2024monkeys}, who show that majority voting plateaus as sample counts grow into the hundreds, our regime is the opposite extreme --- $N = 5$ --- and our finding is about \emph{where} the early gains come from.

\subsection{The confound that inverted a result}

An earlier analysis of our own data showed Qwen3.6-plus on HLE dropping from 14.20\% under the baseline to 11.20\% under \branch{}, which reads as a real and interesting regression: an operator that helps everywhere else actively hurting one model. It was an artifact. The \branch{} run lost 172 of 500 items to read timeouts against the proxy, against 131 for the baseline, and unresolved items were being scored as incorrect. On the 305 items all four Qwen runs completed, \branch{} is $+0.66$. The magnitude of this artifact --- roughly 3.6 points, larger than most true effects in Table~\ref{tab:main} --- is our argument for the protocol in Section~\ref{sec:protocol}. We suspect unpaired scoring is common in multi-method evaluations run against shared or rate-limited endpoints, and that it is rarely reported.

\subsection{Limitations}

Selection in \branch{} is an unweighted majority vote. Verifier- or confidence-weighted selection \citep{cobbe2021verifiers,lightman2024verify} is strictly more expressive, and we record an agreement ratio per item that we never use for weighting; this is the largest gap between our \branch{} and the state of the art. Relatedly, \branch{} is flat parallel sampling, not tree search: it does not expand or prune partial reasoning states as ToT and GoT do \citep{yao2023tot,besta2024got}, so the name describes its position in our operator space rather than its algorithmics. We report no significance tests; a single accuracy figure from a 200-item cell carries a 95\% interval of roughly $\pm 7$ points, and while paired deltas are considerably tighter, we did not quantify them, so small effects in the BBEH and Omni-MATH cells should be read as provisional. Our HLE grader is a lower bound. Omni-MATH was not run on Qwen3.6-plus. Finally, all three models are reasoning models that emit hidden deliberation, and the truncation mechanism of Section~\ref{sec:mechanism} may not transfer to models without that behavior --- although the Qwen3.6-plus cells offer a partial preview of that regime.

\subsection{Why the pilot misled us}

A 50-item pilot preceding the full runs showed \grow{} at $+4$, \prune{} at $+5.3$ on MuSiQue but $-2$ on HLE, and \branch{} at $+8$ --- a pattern of task-dependent operator effectiveness that motivated the routing hypothesis. At full scale these deltas roughly halved and the \prune{} sign flip disappeared. At $n = 50$ the standard error alone is near 7 points, comparable to every effect being measured. We report this because pilots of that size remain common when inference is expensive, and ours would have supported a conclusion the full data contradicts.

\section{Conclusion}

Treating additive refinement, decomposition, and repeated sampling as three operators in one space, and measuring them under identical conditions, gives a clearer picture than the method-at-a-time literature allows. Across 14 model $\times$ benchmark cells, sampling and voting improve accuracy every time, by a mean of six points, while deepening a single path and decomposing into sub-questions are inconsistent and sometimes harmful. The reason sampling wins, however, is not only the one usually given: on long-reasoning models under finite token budgets, much of its benefit is the recovery of answers that a single pass never emitted at all. That points somewhere more specific than ``sample more'' --- namely at the budget-exhaustion failure itself, which is cheaper to attack directly. Our routing hypothesis did not survive contact with the full data, and neither did one of our own intermediate results, which we take as evidence that paired scoring and explicit separation of infrastructure from model failure deserve to be default practice in comparative test-time-compute studies.

\subsubsection*{Reproducibility Statement}

Section~\ref{sec:setup} specifies models, benchmarks, decoding parameters, and token budgets; Section~\ref{sec:protocol} specifies the deduplication, exclusion, and pairing rules used for every number reported. Per-item records --- extracted answer, gold answer, round count, agreement ratio, full trace, and latency --- were retained for all 49{,}327 graded items, so every figure in this paper can be recomputed without re-querying any model. Appendix~\ref{app:failures} reports per-cell attrition. Figures are regenerated from the released numbers by a single script.

\bibliography{references}
\bibliographystyle{iclr2025_conference}

\appendix

\section{Per-Cell Attrition}
\label{app:failures}

Table~\ref{tab:attrition} reports, for each cell, the number of items attempted and the number in the paired set used for scoring. Attrition is the count of items that at least one of the four operators failed to resolve after seven retries with exponential backoff. All failures were transport-level: read timeouts against the proxy (600\,s read timeout), connection resets, and provider-side rate limiting at 30 requests per minute for one model.

\begin{table}[h]
\centering
\caption{Attempted versus paired item counts. Six of the fourteen cells are unaffected. The HLE Qwen3.6-plus cell is the outlier and is the cell discussed in Section~\ref{sec:discussion}.}
\label{tab:attrition}
\small
\begin{tabular}{llrrr}
\toprule
Benchmark & Model & Attempted & Paired $n$ & Attrition \\
\midrule
MuSiQue & DeepSeek-V4-Pro & 2{,}417 & 2{,}417 & 0 \\
MuSiQue & MiniMax-M3      & 2{,}417 & 2{,}398 & 19 \\
MuSiQue & Qwen3.6-plus    & 2{,}417 & 2{,}395 & 22 \\
HLE & DeepSeek-V4-Pro & 2{,}158 & 2{,}152 & 6 \\
HLE & MiniMax-M3      & 500 & 500 & 0 \\
HLE & Qwen3.6-plus    & 500 & 305 & 195 \\
BBEH & DeepSeek-V4-Pro & 200 & 200 & 0 \\
BBEH & MiniMax-M3      & 200 & 200 & 0 \\
BBEH & Qwen3.6-plus    & 200 & 192 & 8 \\
SuperGPQA & DeepSeek-V4-Pro & 300 & 300 & 0 \\
SuperGPQA & MiniMax-M3      & 300 & 299 & 1 \\
SuperGPQA & Qwen3.6-plus    & 300 & 300 & 0 \\
Omni-MATH & DeepSeek-V4-Pro & 300 & 299 & 1 \\
Omni-MATH & MiniMax-M3      & 300 & 282 & 18 \\
\bottomrule
\end{tabular}
\end{table}

\section{Operator Pseudocode}
\label{app:pseudocode}

\begin{algorithm}[h]
\caption{The three recursion operators. $\mathrm{solve}$ is the shared primitive of Section~\ref{sec:method}, including the finalization retry for budget-exhausted generations. $\mathrm{norm}$ is the shared answer normalizer.}
\begin{algorithmic}[1]
\Function{Grow}{$x$, $T = 3$}
  \State $y_0 \gets \mathrm{solve}(x, \varnothing)$
  \For{$t = 1$ to $T-1$}
    \State $y_t \gets \mathrm{solve}(x, y_{t-1})$
    \If{$\mathrm{norm}(y_t) = \mathrm{norm}(y_{t-1})$} \Return $y_t$ \Comment{answer stable}
    \EndIf
  \EndFor
  \State \Return $y_{T-1}$
\EndFunction
\Statex
\Function{Prune}{$x$}
  \State $q_1,\dots,q_k \gets \mathrm{decompose}(x)$
  \If{$k = 1$} \Return $\mathrm{solve}(x, \varnothing)$ \Comment{adaptive fallback}
  \EndIf
  \For{$i = 1$ to $k$}
    \State $a_i \gets \mathrm{solve}(q_i, \{(q_j, a_j)\}_{j < i})$ \Comment{later hops see earlier answers}
  \EndFor
  \State \Return $\mathrm{compose}(x, \{(q_i, a_i)\})$
\EndFunction
\Statex
\Function{Branch}{$x$, $N = 5$, $\tau = 0.7$}
  \State $s_1,\dots,s_N \gets$ $N$ independent draws of $\mathrm{solve}(x, \varnothing)$ at temperature $\tau$
  \State \Return $\arg\max_{k} \; |\{ i : \mathrm{norm}(s_i) = k \}|$ \Comment{empty answers normalize to a null key}
\EndFunction
\end{algorithmic}
\end{algorithm}

\end{document}